%% file: main.tex
\documentclass[10pt,twocolumn,letterpaper]{article}
\usepackage[table]{xcolor}
\usepackage[pagenumbers]{arXiv}
\usepackage{xr-hyper}

\usepackage[pagebackref,breaklinks,colorlinks]{hyperref}
\usepackage[capitalize]{cleveref}

\usepackage{xspace}
\usepackage{amsmath}
\usepackage{amssymb}
\usepackage{xcolor}
\usepackage{booktabs}
\usepackage{multirow}
\usepackage{pifont}
\usepackage{tabularx}
\usepackage{algpseudocode}
\usepackage{amsfonts}
\usepackage{hhline}
\usepackage{algorithm}
\usepackage{marvosym}
\usepackage{textcomp}

\newcommand{\method}{HERO\xspace}

\begin{document}
\title{\method: \underline{H}ierarchical \underline{E}xtrapolation and \underline{R}efresh for Efficient W\underline{o}rld Models}

\author{
    Quanjian Song \quad
    Xinyu Wang \quad
    Donghao Zhou \quad
    Jingyu Lin \\
    Cunjian Chen \quad
    Yue Ma
}

\twocolumn[{%
\renewcommand\twocolumn[1][]{#1}%
\maketitle
\input{sec/overview}
}]

\renewcommand{\thefootnote}{\Letter}
\footnotetext[1]{Corresponding author.}

\input{sec/0_abstract}
\input{sec/1_intro}
\input{sec/2_related_works}

\input{sec/3_method}

\input{sec/4_exp}

\input{sec/5_conclusion}

{\small
\bibliographystyle{ieeenat_fullname}
\bibliography{references}
}

\clearpage
\appendix
\input{sec/X_suppl}

\end{document}

%% file: sec/overview.tex
\begin{center}
\centering
\vspace{-5.5mm}
\includegraphics[width=1.\textwidth]{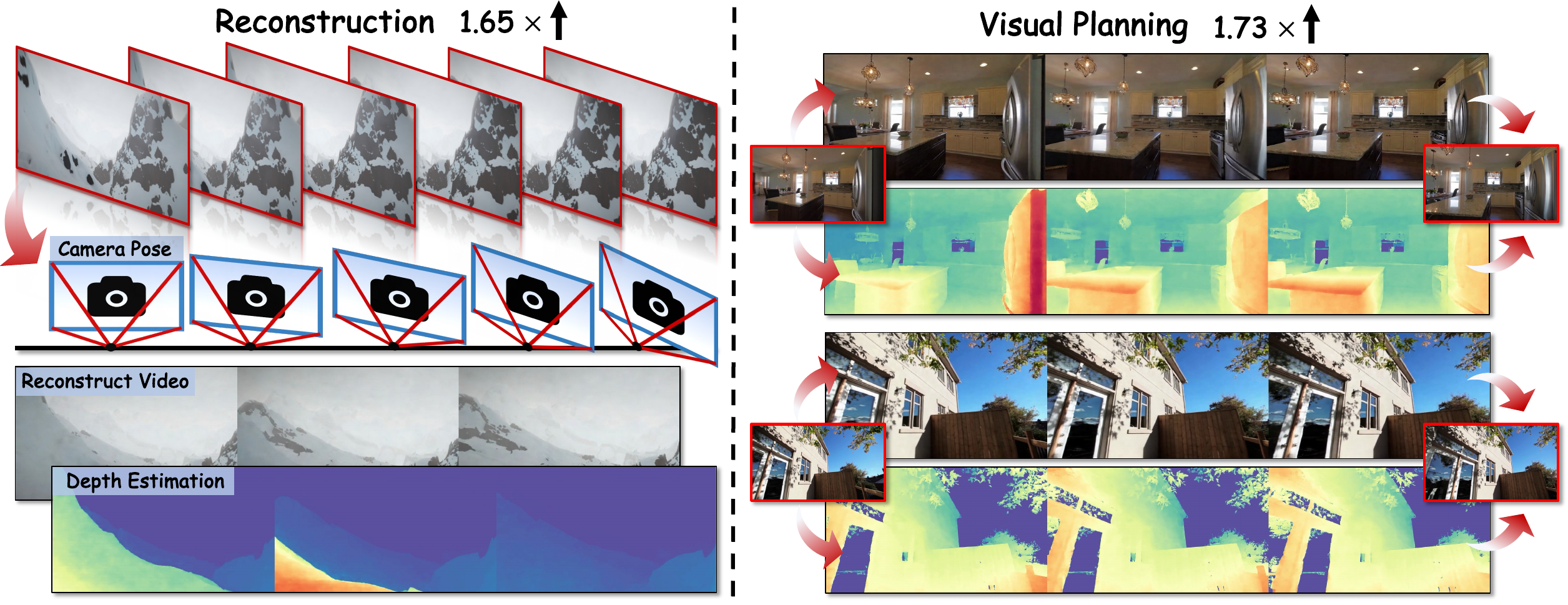}
\vspace{-4.5mm}
\captionsetup{hypcap=false}
\captionof{figure}{ 
Showcase of our \method, which accelerates world model frameworks like Aether with minimal quality degradation. 
}
\label{fig:overview}
\end{center}

%% file: sec/0_abstract.tex
\begin{abstract}
Generation-driven world models create immersive virtual environments but suffer slow inference due to the iterative nature of diffusion models.
While recent advances have improved diffusion model efficiency, directly applying these techniques to world models introduces limitations such as quality degradation.
In this paper, we present \method, a training-free hierarchical acceleration framework tailored for efficient world models.
Owing to the multi-modal nature of world models, we identify a feature coupling phenomenon, wherein shallow layers exhibit high temporal variability, while deeper layers yield more stable feature representations.
Motivated by this, \method adopts hierarchical strategies to accelerate inference:
(i) In shallow layers, a patch-wise refresh mechanism efficiently selects tokens for recomputation. With patch-wise sampling and frequency-aware tracking, it avoids extra metric computation and remain compatible with FlashAttention.
(ii) In deeper layers, a linear extrapolation scheme directly estimates intermediate features. This completely bypasses the computations in attention modules and feed-forward networks.
Our experiments show that \method achieves a 1.73$\times$ speedup with minimal quality degradation, significantly outperforming existing diffusion acceleration methods.
\end{abstract}

%% file: sec/1_intro.tex
\section{Introduction}
\begin{quote}
    \textit{``World models are the key technical pathways toward Artificial General Intelligence." }
    \begin{flushright}
        -- Yann LeCun
    \end{flushright}
\end{quote}
The essence of artificial intelligence lies in the pursuit of Artificial General Intelligence (AGI) systems~\cite{hafner2023mastering}, 
which are expected to perceive, reason, and act across a wide range of tasks with human-like adaptability.
While large language models (LLMs)~\cite{GPT-4,Qwen,LLaMa,LLaMa2,gpt} were once widely regarded as the leading candidates for achieving AGI, an emerging perspective~\cite{wu2023daydreamer,Sora,navigation_world_models} now views world models as a more fundamental and promising stepping stone toward this goal.
Among various world models, generation-driven approaches~\cite{agarwal2025cosmos,russell2025gaia,bar2025navigation,Aether} have recently gained increasing attention for their strong performance. 
Based on powerful video generators, these world models aim to capture the spatial content and temporal motion of the real world from large-scale data. They enable the synthesis of immersive, coherent, and interactive virtual environments.
However, these world models still have a large challenge about inference speed due to the inefficient diffusion transformer,  limiting their practicality in real-time potential application.

\begin{figure}[!t]
\centering
\includegraphics[width=1.0\columnwidth]{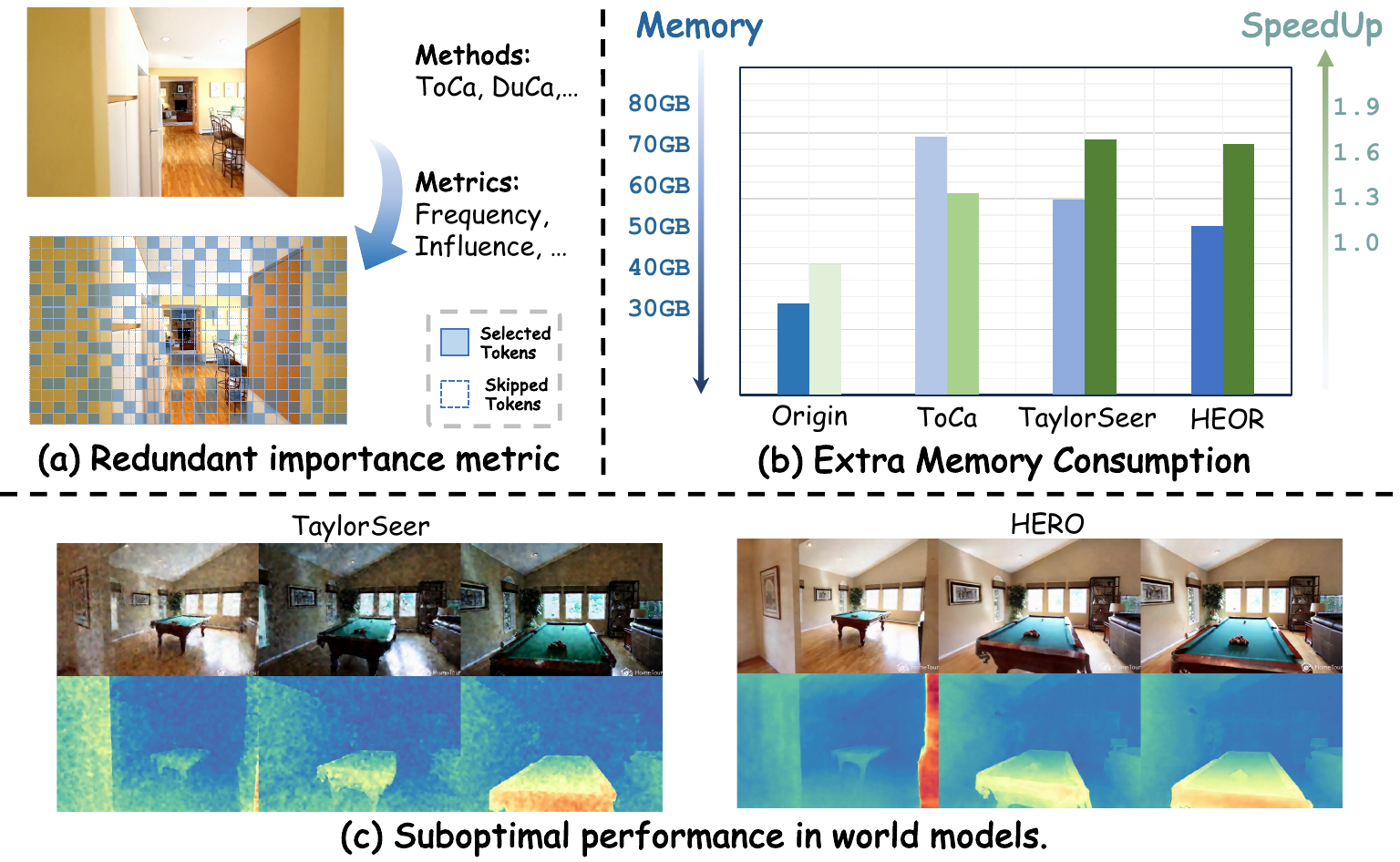} %
\caption{
Applying diffusion-based acceleration methods to world models directly exposes several limitations:
(a) Redundant importance metric.
(b) Extra memory consumption.
(c) Suboptimal performance in world models.
}
\label{fig:bad_case}
\end{figure}

Recently, various techniques~\cite{yan2024perflow,kim2024token,zhang2024effortless,wu2025quantcache} have been proposed to improve diffusion model efficiency.
Among them, cache-based approaches~\cite{FoRA,ToCa,TaylorSeer} achieve efficiency gains by reducing cost through feature reuse or forecasting.
A naive way is to apply these accelerate methods directly to world models.
However, as illustrated in Figure\,\ref{fig:bad_case}, these approaches faces three key limitations in the context of world models:
\textit{(i) Redundant importance metric.}
Most methods use importance metrics to decide which tokens to recompute, such as attention scores in ToCa~\cite{ToCa} and matrix norms in DuCa~\cite{DuCa}.
However, these metrics lack theoretical guarantees, which can lead to inaccurate token selection, and their computation introduces additional overhead.
\textit{(ii) Extra memory consumption.}
Several methods like ToCa~\cite{ToCa} stores attention scores during computation to select important tokens for recomputation.
However, this is incompatible with FlashAttention~\cite{FlashAttention} and can easily cause out-of-memory issues, especially in video tasks.
\textit{(iii) Suboptimal performance in world models.}
\textbf{Importantly}, as shown in Figure~\ref{fig:bad_case}(c), directly applying existing diffusion acceleration methods, such as TaylorSeer~\cite{TaylorSeer}, to world models leads to poor generation quality.
Thus, a question emerges: \textit{Why do general acceleration methods fail on world models, even when they share the same base model?}
%

Using Aether~\cite{Aether} as a representative example, we identify two distinctive characteristics of world models:
\textit{(i) Feature coupling in world models.}
Unlike traditional video generators, world models take multi-modal (e.g., depth maps, camera poses) inputs and outputs, which are concatenated along the channel dimension and mapped into coupled tokens.
Due to varying modality sensitivities, coupled tokens behave differently across MMDiT layers.
Yet, existing diffusion-based acceleration methods apply a uniform strategy across all layers, ignoring the inherently hierarchical structure of world models.
\textit{(ii) Hierarchical patterns in MMDiTs.}
Due to significant differences among input modalities, channel-wise concatenation amplifies these discrepancies and tends to propagate errors into deeper layers.
In contrast, iterative processing across layers gradually mitigates modality differences, resulting in more stable deep-layer features that are also less sensitive to errors.

In this paper, we propose \textbf{\method}, a training-free hierarchical acceleration method for efficient world models.
Our analysis reveals that shallower MMDiT layers exhibit higher temporal variability, whereas deeper MMDiT layers yield more stable feature representations.
Motivated by this, \method adopts hierarchical strategies:
\textit{(i) In the shallow layers}, we employ a \textbf{patch-wise refresh} strategy that dynamically selects tokens to recompute or reuse.
Leveraging local similarity within videos, we divide tokens into non-overlapping patches and randomly select a subset in each patch for recomputation, avoiding additional metric computation while remaining compatible with FlashAttention~\cite{FlashAttention}.
In addition, a frequency-aware tracking mechanism is incorporated to mitigate error accumulation.
\textit{(ii) In the deeper layers}, we adopt a \textbf{linear extrapolation} scheme to directly estimate intermediate features, thereby skipping redundant computations of the full attention and feed-forward network.
Extensive experiments show that \method delivers a 1.73$\times$ speedup while preserving high performance on both visual planning and reconstruction, outperforming existing diffusion acceleration methods.


%% file: sec/2_related_works.tex
\section{Related Works}

\begin{figure*}[!t]
\centering
\includegraphics[width=0.98\textwidth]{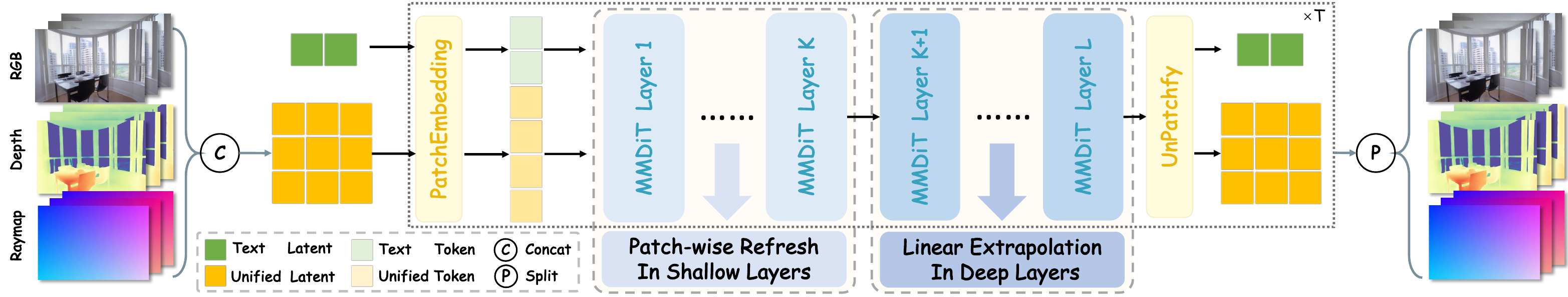} %
\caption{
Overall workflow of \method, with hierarchical acceleration strategies in world model Aether: patch-wise refresh in shallow layers for dynamic token recomputation, and linear extrapolation in deep layers to skip intermediate computations.
}
\label{fig:overall_framework}
\end{figure*}

\noindent
\textbf{Efficient Diffusion Models.}
Despite the success of diffusion models, their inference remains costly, prompting two main acceleration methods:
\textit{(i) Optimized-based Acceleration.}
Early works~\cite{song2020denoising,lu2022dpm} accelerate inference through deterministic sampling and high-order solvers, while recent methods~\cite{yan2024perflow,yin2024one,luo2024one} further refine ODE trajectories via distillation, enabling fewer denoising steps.
\textit{(ii) Structure-based Acceleration.}
Early works reduce computational overhead by simplifying model architectures through token compression~\cite{kim2024token, zhang2025training}, pruning~\cite{fang2023structural, zhang2024effortless}, and quantization~\cite{shang2023post, wu2025quantcache}.
Recently, caching-based methods such as FoRA~\cite{FoRA}, ToCa~\cite{ToCa}, and TaylorSeer~\cite{TaylorSeer} demonstrate notable success by reusing or predicting intermediate features across steps to reducing redundant computation.
However, directly applying these methods to world models degrades generation quality, as their uniform layer-wise strategy conflicts with the hierarchical nature of world models.
This motivates \method, which introduces hierarchical strategies for effective acceleration.

\noindent
\textbf{Unified World Models.}
The unified world model aims to simulate world dynamics and enable reasoning, planning, and decision-making. 
Existing methods fall into two main categories.
\textit{(i) Policy-driven models.}
These models~\cite{hafner2019dream, wu2023daydreamer, hafner2023mastering, robine2023transformer} rely on agents that learn about the world through interaction with the environment.
By performing actions and receiving observations and rewards, the agent gradually builds and refines an internal representation of the world.
The core challenge lies in learning a compact yet expressive world model from experience and using it to guide future decisions.
\textit{(ii) Generation-driven models.}
These models~\cite{agarwal2025cosmos, Aether, russell2025gaia, bar2025navigation} aim to capture the spatio-temporal dynamics of the world using large-scale datasets, enabling them to generate realistic, coherent, and interactive virtual environments.
Aether~\cite{Aether}, as a representative example, is fine-tuned from a video foundation model and supports both multimodal input and output, closely mirroring real-world perception.
In this paper, we focus on generation-driven models, which achieve strong performance but often suffer from slow inference due to the iterative nature of diffusion models.
This limitation motivates our work: exploring efficient generation-driven world models.


%% file: sec/3_method.tex
\section{Efficient World Models}

%
In this section, we propose \method, a training-free hierarchical acceleration method for efficient world models.
Using Aether as an example, we first analyze the architecture of recent world models, where multi-modal inputs and outputs introduce feature coupling within MMDiTs.
We then conduct a hierarchical feature analysis, which reveals that shallow layers exhibit greater variability, while deeper layers remain more stable.
Building on this insight, \method introduces hierarchical strategies: patch-wise refresh in shallow layers for dynamic token recomputation, and linear extrapolation in deep layers to skip intermediate computations.
Figure\,\ref{fig:overall_framework} illustrates the overall workflow of our \method.

\subsection{Framework of World Models}
\label{sec:4.1}
Existing world models extend DiT-based video generators (e.g., Wan~\cite{Wan}, HunyuanVideo~\cite{HunyuanVideo}, CogVideoX~\cite{CogvideoX}) with multi-modal inputs and outputs.
Herein, We illustrate this design using Aether~\cite{Aether}, a recent SOTA method built upon CogVideoX, and operate in latent space for simplicity.

In addition to video latent $Z_v \in \mathbb{R}^{f \times h \times w \times c_v}$, Aether incorporates depth latent $Z_d \in \mathbb{R}^{f \times h \times w \times c_d}$ and camera latent $Z_c \in \mathbb{R}^{f \times h \times w \times c_c}$, all sharing the same spatial-temporal size but different channel dimensions $(c_d, c_c)$.
All latents are concatenated along the channel, as formulated by:
\begin{equation}
    Z = \operatorname{Concat}([Z_v, Z_d, Z_c], dim=-1),
\label{eq:0}
\end{equation}
where $Z \in \mathbb{R}^{f \times h \times w \times (c_v + c_d + c_c) }$. Finally, the unified latent is jointly fed into the network with the encoded text latent $\mathcal{T} \in \mathbb{R}^{N \times c_{t}}$, and $c_{t}$ is the text embedding dimension.

During each forward pass, the unified latent $Z$ and text latent $\mathcal{T}$ are first passed through PatchEmbedding, producing unified tokens $z \in \mathbb{R}^{(f \cdot h' \cdot w') \times d}$ and text tokens $\tau \in \mathbb{R}^{N \times d}$, where $d$ is the unified feature dimension.
These tokens are then processed by a stack of $L=42$ MMDiT layers. An unpatchify operation subsequently restores the output to its original spatial shape: $\bar{Z} \in \mathbb{R}^{f \times h \times w \times (c_v + c_d + c_c)}$.
Finally, the disentangled unified latent $\bar{Z}$ is split along channel dimension to recover three modality $\bar{Z_v}$, $\bar{Z_d}$ and $\bar{Z_c}$:
\begin{equation}
    \bar{Z_v}, \bar{Z_d}, \bar{Z_c} = \operatorname{Split}(\bar{Z}, dim=-1).
\label{eq:1}
\end{equation}

The multi-modal fusion in Eq.\,(\ref{eq:0}) and disentanglement in Eq.\,(\ref{eq:1}) reveal a distinctive property of world models: \textit{The features of MMDiTs are strongly coupled and hard to disentangle.} This motivates the hierarchical analysis below.

\begin{figure*}[t]
\centering
\includegraphics[width=0.98\textwidth]{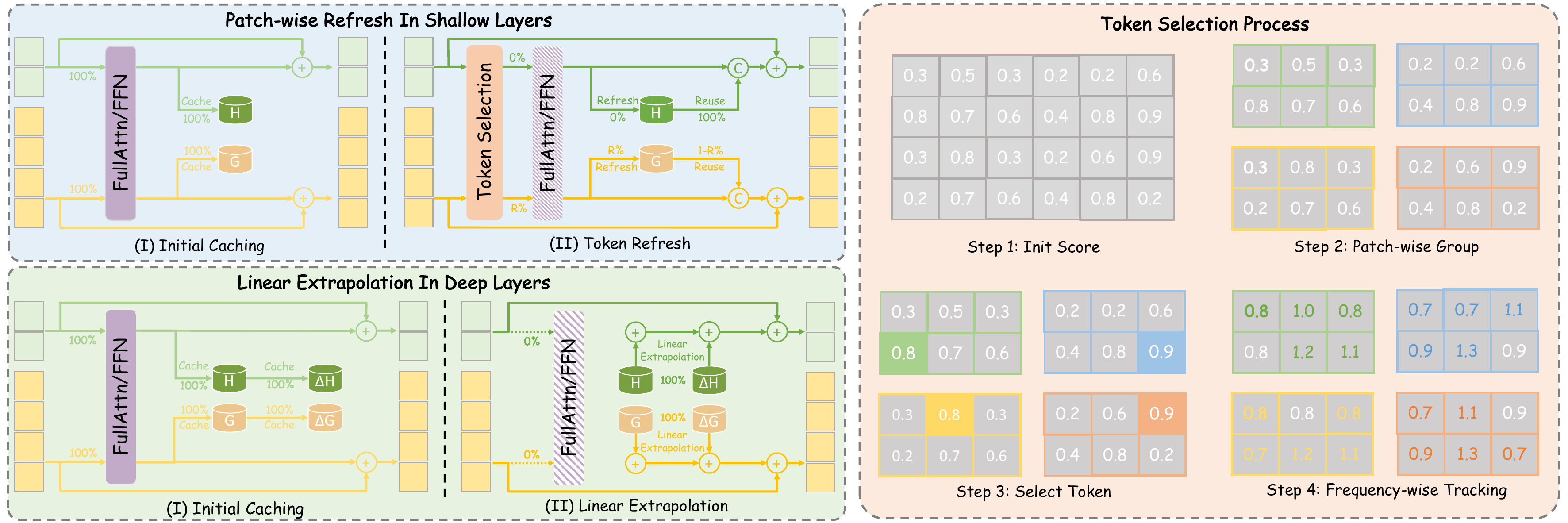} %
\caption{
Toy examples of our hierarchical strategy: patch-wise refresh in shallow layers and linear extrapolation in deep layers.
}
\label{fig:method}
\end{figure*}

\subsection{Hierarchical Analysis of World Models}
\label{sec:4.2}
We begin with a brief overview of the MMDiT layers in the Aether (omitting scale and shift for clarity) and existing caching strategies, then analyze the limitations of these approaches in world models from a hierarchical perspective.
%

\noindent
\textbf{Caching in MMDiTs.}
For each timestep $t \in [1, T]$ and layer $l \in [1, L]$ in MMDiT, the unified input tokens $z_t^l$ and text tokens $\tau_t^l$ jointly undergo a series of transformations.
Each transformation applies LayerNorm, then function $\mathcal{F}(\cdot)$ (FullAttn or FFN), producing intermediate features $\mathcal{G}_t^l$ and $\mathcal{H}_t^l$.
These features are added to the original inputs $z_t^{l}$ and $\tau_t^{l}$,  yielding the updated unified tokens and text tokens:
\begin{equation}
\begin{aligned}
    \mathcal{G}_t^{l}, \: \mathcal{H}_{t}^{l} &= \mathcal{F}(z_t^{l}, \tau_t^{l}), \\
    z_{t}^{l}, \: \tau_{t}^{l} &\leftarrow z_{t}^{l} + \mathcal{G}_{i}^{l}, \: \tau_{t}^{l} + \mathcal{H}_i^{l}.
\end{aligned}
\label{eq:2}
\end{equation}
%

Owing to the iterative nature of diffusion models, features across adjacent timesteps are often highly similar.
Leveraging this property, recent caching-based methods~\cite{FoRA,ToCa,TaylorSeer} propose various strategies to cache and reuse the intermediate features $\mathcal{G}$ and $\mathcal{H}$ across all layers for efficient inference.
However, as shown in Figure\,\ref{fig:bad_case}(c), directly applying techniques like TaylorSeer~\cite{TaylorSeer} to Aether leads to degraded quality in the generated videos, depth maps, and raymaps.
This observation motivates a hierarchical analysis of MMDiT in the context of world models.

\begin{figure}[!t]
\centering
\includegraphics[width=0.95\columnwidth]{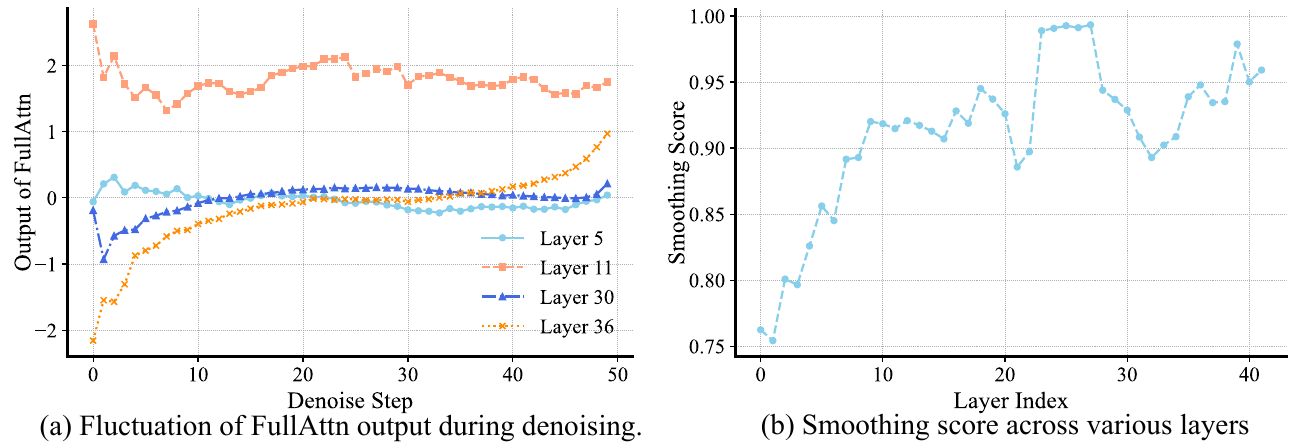} %
\caption{
Hierarchical feature analysis: shallow layers show greater variability and benefit from recomputation, while deep layers are more stable and better suited for reuse.
}
\label{fig:analysis}
\end{figure}

\noindent
\textbf{Hierarchical Feature Analysis.}
Recall Eq.\,(\ref{eq:0}) and Eq.\,(\ref{eq:1}) in previous section, where the multi-modal fusion of the world model introduces feature coupling in MMDiTs.
Given the varying sensitivities of different modalities at each layer, we hypothesize that \textit{coupled features exhibit distinct characteristics across MMDiT layers}.
This hypothesis is able to explain why existing methods often yield suboptimal results, as they typically apply a uniform strategy across all layers without fully considering these differences.

To evaluate this hypothesis, we analyze the evolution of features from different layers during the denoising process.
As shown in Figure\,\ref{fig:analysis}(a), features from layer $5$ and $11$ exhibit notable fluctuations, while those from layer $30$ and $36$ remain relatively stable.
To quantify this, we compute the variance of the second-order differences of each layer’s temporal features and normalize results to the range $[0, 1]$, where higher values indicate greater stability.
In Figure\,\ref{fig:analysis}, shallow layers yield lower scores, reflecting greater instability, while deeper layers show higher scores, indicating more stable representations.
We attribute this phenomenon to modality differences amplified by channel-wise concatenation, leading to instability in shallow layers.
As layers progress, features are gradually assimilated, reducing modality differences and yielding stable representations.

Therefore, we can conclude that: shallow MMDiT layers exhibit greater variability and benefit from feature recomputation, while deeper layers are more stable and better suited for reuse.
Based on this insight, we devise hierarchical acceleration strategies for shallow and deep layers to exploit their distinct characteristics.
Toy examples are illustrated in Figure\,\ref{fig:method}, with the
pseudocode provided in Algorithm\,\ref{alg:1}.

\begin{algorithm}[!t]
\footnotesize
\caption{\textit{Hierarchical Extrapolation and Refresh}}
\textbf{Require:} The timestep sequences $\{t, t - 1, \ldots, t - M\}$, the timestep interval $M$, the select function $\mathcal{S}(\cdot)$.
\begin{algorithmic}[1]
\State \textit{// Initial Caching.}
\For {$l=1, \dots, L$}
    \For{$\mathcal{F}_i \in \{\text{FullAttn}_l, \text{FFN}_l$\}}
        \State $\mathcal{G}_t^{l}, \mathcal{H}_t^{l} = \mathcal{F}_i(z_t^{l}, \tau_t^{l})$
        \State $\Delta \mathcal{G}_t^{l}, \Delta \mathcal{H}_t^{l} \gets \mathcal{G}_t^{l} - \mathcal{G}_{t + M}^{l}, \mathcal{H}_t^{l} - \mathcal{H}_{t + M}^{l}$
        \State $z_t^{l}, \tau_t^{l} \gets z_t^{l} + \mathcal{G}_t^{l}, \tau_t^{l} {+} \mathcal{H}_t^{l}$
    \EndFor
\EndFor
\State \textit{// Hierarchical Extrapolation and Refresh.}
\For{$k = 1, 2, \dots, M$}
    \For {$l=1, \dots, L$}
        \For{$\mathcal{F}_i \in \{\operatorname{FullAttn}_l, \operatorname{FFN}_l$\}}
            \If {$l < K$}
                \State $\mathcal{G}_{t-k}^{l}, \mathcal{H}_{t-k}^{l} = \mathcal{F}_i(\mathcal{S}(z_{t-k}^{l}), \mathcal{S}(\tau_{t-k}^{l}))$
                \State $\mathcal{G}_{t-k}^{l} \gets \mathcal{G}_{t-k}^{l} \cup \mathcal{G}_{t}^{l}[z_{t-k}^{l} \setminus \mathcal{S}(z_{t-k}^{l})]$
                \State $\mathcal{H}_{t-k}^{l} \gets \mathcal{H}_{t-k}^{l} \cup \mathcal{H}_{t}^{l}[\tau_{t-k}^{l} \setminus \mathcal{S}(\tau_{t-k}^{l})]$
            \Else
                \State $\mathcal{G}_{t-k}^{l} = \mathcal{G}_k^{l} + \frac{\Delta \mathcal{G}_t^{l}}{M} \cdot k$
                \State $\mathcal{H}_{t-k}^{l} = \mathcal{H}_k^{l} + \frac{\Delta \mathcal{H}_t^{l}}{M} \cdot k$
            \EndIf
            \State $z_{t-k}^{l}, \tau_{t-k}^{l} \gets z_{t-k}^{l} + \mathcal{G}_{t-k}^{l}, \tau_{t-k}^{l} {+} \mathcal{H}_{t-k}^{l}$
        \EndFor
    \EndFor
\EndFor
\end{algorithmic}
\label{alg:1}
\end{algorithm}

\subsection{Patch-wise Refresh in Shallow Layers}
\label{sec:4.3}
As analyzed in the previous section, shallow-layer features are unstable and prone to error amplification, making direct reuse unreliable.
Motivated by this, we employ a patch-wise refresh strategy in the first $K$ layers of MMDiT, enabling dynamic token refreshing.
For each layer $l \in [1, K)$, we consider timesteps $\{t, t - 1, \ldots, t - M\}$ with interval $M$ to perform initial caching and subsequent token refreshing.

At timestep $t$, the unified tokens $z_t^{l}$ and text tokens $\tau_t^{l}$ are sequentially processed by Eq.\,(\ref{eq:2}). The resulting intermediate features $\mathcal{G}_t^{l}$ and $\mathcal{H}_t^{l}$, will be cached for subsequent refreshing.

At each subsequent timestep $t - k$ where $k \in [1, M]$, the unified tokens $z_{t-k}^l$ and text tokens $\tau_{t-k}^l$ are filtered by the selection function $\mathcal{S}$.
The selected tokens $\mathcal{S}(z_{t-k}^l)$ and $\mathcal{S}(\tau_{t-k}^l)$, are then refreshed using Eq.\,(\ref{eq:2}).
While the remaining tokens $z_{t-k}^{l} \setminus \mathcal{S}(z_{t-k}^{l})$ and $\tau_{t-k}^{l} \setminus \mathcal{S}(\tau_{t-k}^{l})$, directly reuse the cached features $\mathcal{G}_t^{l}$ and $\mathcal{H}_t^{l}$ from timestep $t$.
Finally, the two branches are merged to update intermediate features:
%
%
\begin{equation}
\begin{aligned}
    \mathcal{G}_{t-k}^{l}, \mathcal{H}_{t-k}^{l} &= \mathcal{F}(\mathcal{S}(z_{t-k}^{l}), \mathcal{S}(\tau_{t-k}^l)),
    \\
    \mathcal{G}_{t-k}^{l} &\leftarrow \mathcal{G}_{t-k}^{l} \cup \mathcal{G}_{t}^{l}[z_{t-k}^{l} \setminus \mathcal{S}(z_{t-k}^{l})],
    \\
    \mathcal{H}_{t-k}^{l} &\leftarrow \mathcal{H}_{t-k}^{l} \cup \mathcal{H}_{t}^{l} [ \tau_{t-k}^{l} \setminus \mathcal{S}(\tau_{t-k}^{l})].
\end{aligned}
\end{equation}

We next detail how the selection function $\mathcal{S}$ identifies tokens to be refreshed, with the toy example is provided in Figure\,\ref{fig:method}.
Notably, this process eliminates explicit importance metrics, thereby reducing computational overhead and ensuring compatibility with FlashAttention~\cite{FlashAttention}.

\begin{table*}[!t]
\centering
\resizebox{\linewidth}{!}{  
\begin{tabular}{lccccccccc}
\toprule
Methods & Subj. Cons. $\uparrow$ & Back. Cons. $\uparrow$ & Mot. Smooth. $\uparrow$ & Dyn. Deg. $\uparrow$ & Aest. Qual. $\uparrow$ & Img. Qual. $\uparrow$ & Latency(s) $\downarrow$ & FLOPs(T) $\downarrow$ & Speed $\uparrow$ \\
\toprule
CogVideoX-I2V             & 92.72 & 93.87 & 98.94 & 61.26 & 50.79 & 68.64 & - & - & - \\
\cmidrule{1-10}
Aether                    & 93.70 & 95.21 & 98.03 & 90.09 & 49.33 & 64.21 & 191.41 & 14352.50 & 1.00$\times$  \\
Aether (FoRA, $M$=2)             & 91.94($\Delta$=-1.76) & 94.53($\Delta$=-0.68) & 97.58($\Delta$=-0.18) & \textbf{90.09}($\Delta$=0.00) & 47.62($\Delta$=-1.71) & 61.90($\Delta$=-2.31) & 108.43 & 7753.18  & 1.85$\times$ \\
Aether (ToCa, $M$=2)             & 92.03($\Delta$=-1.67) & 94.37($\Delta$=-0.84) & \underline{97.86}($\Delta$=-0.17) & \textbf{90.09}($\Delta$=0.00) & \underline{49.68}($\Delta$=+0.35) & 60.76($\Delta$=-3.45) & 138.45 & 10053.38 & 1.43$\times$ \\
Aether (TaylorSeer, $M$=2)       & 89.06($\Delta$=-4.64) & 93.62($\Delta$=-1.59) & 96.23($\Delta$=-1.80) & \underline{89.18}($\Delta$=-0.91) & 47.78($\Delta$=-1.55) & 58.55($\Delta$=-5.66)  & 108.43 & 7753.18 & 1.85$\times$ \\
\rowcolor{gray!25}
Aether (\method, $M$=2)             & \textbf{93.42}($\Delta$=-0.28) & \textbf{95.02}($\Delta$=-0.19) & \textbf{97.90}($\Delta$=-0.13) & \textbf{90.09}($\Delta$=0.00) & 49.34($\Delta$=+0.01) & \textbf{64.21}($\Delta$=0.00) & \underline{113.98} & \underline{8295.21} & \underline{1.73$\times$} \\
\rowcolor{gray!25}
Aether (\method, $M$=3)             & \underline{92.05}($\Delta$=-1.65) & \underline{94.90}($\Delta$=-0.31) & 97.53($\Delta$=-0.50) & \textbf{90.09}($\Delta$=0.00) & \textbf{50.02}$\Delta$=+0.69 & \underline{61.92}($\Delta$=-2.29)  & \textbf{101.13} & \textbf{7241.78} & \textbf{1.97$\times$} \\
\bottomrule
\end{tabular}
}
\caption{
Comparison of performance and efficiency with existing methods in visual planning task.
All performance metrics are reported on VBench~\cite{VBench}. Best scores are in \textbf{bold} and second-best are \underline{underlined}, among acceleration methods.
}
\label{tab:eval_video}
\end{table*}

\begin{table*}[!t]
\centering
\resizebox{\linewidth}{!}{  
\begin{tabular}{l ccc ccc ccc}
\toprule
\multirow{2}{*}{Methods} & \multicolumn{3}{c}{Video Depth Estimation} & \multicolumn{3}{c}{Camera Pose Estimation} & \multicolumn{3}{c}{Efficiency} \\
\cmidrule(lr){2-4} \cmidrule(lr){5-7} \cmidrule(lr){8-10}
 & Abs Rel $\downarrow$ & $\delta < 1.25$ $\uparrow$ & $\delta < 1.25^2$ $\uparrow$ & ATE $\downarrow$ & RPE trans $\downarrow$ & RPE rot $\downarrow$ & Latency(s) $\downarrow$ & FLOPs(T) $\downarrow$ & Speed $\uparrow$ \\
\toprule
DUS3TR               & 0.656 & 0.452 & - & 0.290 & 0.132 & 7.869 & - & - & - \\
CUT3R                & 0.421 & 0.479 & - & 0.213 & 0.066 & 0.621 & - & - & - \\
\cmidrule{1-10}
Aether               & 0.340 & 0.502 & 0.738 & 0.177 & 0.068 & 0.780 & 55.42 & 4305.60 & 1.00$\times$ \\
Aether (FoRA, $M$=2)        & \underline{0.345}($\Delta$=-0.005) & 0.483($\Delta$=-0.019) & \underline{0.733}($\Delta$=-0.005) & 0.193($\Delta$=-0.016) & 0.084($\Delta$=-0.016) & 0.891($\Delta$=-0.111) & 30.45 & 2440.64 & 1.76$\times$ \\
Aether (ToCa, $M$=2)        & 0.349($\Delta$=-0.009) & 0.488($\Delta$=-0.014) & 0.716($\Delta$=-0.022) & 0.217($\Delta$=-0.040) & \underline{0.071}($\Delta$=-0.003) & 0.862($\Delta$=-0.082) & 39.63 & 3373.24 & 1.28$\times$ \\
Aether (TaylorSeer, $M$=2)  & 0.356($\Delta$=-0.016) & 0.466($\Delta$=-0.036) & 0.728($\Delta$=-0.010) & 0.205($\Delta$=-0.028) & 0.087($\Delta$=-0.019) & 1.105($\Delta$=-0.325) & 30.45 & 2440.64 & 1.76$\times$ \\
\rowcolor{gray!25}
Aether (\method, $M$=2)        & \textbf{0.344}($\Delta$=-0.004) & \textbf{0.502}($\Delta$=0.000) & \textbf{0.734}($\Delta$=-0.004) & \underline{0.182}($\Delta$=-0.005) & \textbf{0.069}($\Delta$=-0.001) & \textbf{0.816}($\Delta$=-0.036) & \underline{32.34} & \underline{2593.90} & \underline{1.65$\times$} \\
\rowcolor{gray!25}
Aether (\method, $M$=3)        & 0.347($\Delta$=-0.007) & \underline{0.490}($\Delta$=0.012) & 0.716($\Delta$=0.022) & \textbf{0.181}($\Delta$=-0.004) & \underline{0.071}($\Delta$=-0.003) & \underline{0.861}($\Delta$=-0.081) & \textbf{27.44} & \textbf{2198.88} & \textbf{1.96$\times$} \\
\bottomrule
\end{tabular}
}
\caption{
Comparison of performance and efficiency with existing methods in reconstruction task.
Best scores are in \textbf{bold} and second-best are \underline{underlined}, among various acceleration methods.%
}
\label{tab:eval_recons}
\end{table*}

\noindent
\textbf{Patch-wise Sampling.}
Nearby frames and regions in videos often share similar features, and this inherent similarity leads to significant redundancy.
To leverage this property, we start by dividing the unified token $z \in \mathbb{R}^{f \times h' \times w' \times d}$ into $P = f \cdot h' \cdot w' / (p_h \cdot p_w)$ non-overlapping patches $\{\mathcal{P}^i\}_{i=1}^P$, where each $\mathcal{P}^i \in \mathbb{R}^{p_h \times p_w \times d}$.
Since tokens within each patch are often similar, we sample a fixed ratio $R$ of representative tokens from each patch. All sampled tokens are then aggregated to form the final selection.
In summary, the selection function $\mathcal{S}$ can be defined as follows:
\begin{equation}
    \mathcal{S}(z) := \bigcup_{i=1}^{P} \operatorname{Sample}(\mathcal{P}^i, R), \quad \text{where} \: \bigcup_{i=1}^{P} \mathcal{P}^{i} = z.
\end{equation}
As for text tokens $\tau$, since they are much fewer than unified tokens $z$, we skip their refresh and bypass the computation.

\noindent
\textbf{Patch-wise Sampling.}
In practice, the stochastic nature of sampling may leave some tokens unrefreshed over time, leading to error accumulation and degraded output quality.
To mitigate this, as shown in Figure\,\ref{fig:method}, we track how long each token has remained unsampled and assign higher sampling probabilities to those neglected longer.

\subsection{Linear Extrapolation in Deep Layers}
\label{sec:4.4}
Recall previous conclusion that, deep-layer features tend to be stable over time, exhibiting predictable patterns.
Therefore, we adopt a linear extrapolation strategy starting from the $k$-th layer of MMDiT, thereby skipping computations of $\operatorname{FullAttn}$ and $\operatorname{FFN}$.
For each layer $l \in [K, L)$, we consider timesteps $\{t, t - 1, \ldots, t - M\}$ with interval $M$ to perform initial caching and subsequent feature extrapolation.

At timestep $t$, the unified tokens $z_t^{l}$ and text tokens $\tau_t^{l}$ are sequentially processed by Eq.\,(\ref{eq:2}).
Unlike in shallow layers, we cache not only the intermediate features $\mathcal{G}_t^{l}$ and $\mathcal{H}_t^{l}$, but also their temporal differences, defined as $\Delta \mathcal{G}_t^{l} = \mathcal{G}_t^l - \mathcal{G}_{t+M}^{l}$ and $\Delta \mathcal{H}_t^{l} = \mathcal{H}t^l - \mathcal{H}_{t+M}^{l}$, in deeper layers.
These additional signals provide useful context for subsequent extrapolation.

At each subsequent timestep $t - k$, where $k \in [1, M]$, we directly estimate intermediate features $\mathcal{G}_{t-k}^{l}$ and $\mathcal{H}_{t-k}^{l}$ using cached features $\mathcal{G}_t^{l}$ and $\mathcal{H}_t^{l}$, along with their differences $\Delta \mathcal{G}_t^{l}$ and $\Delta \mathcal{H}_t^{l}$.
Given the stability and predictability of deep-layer features, we apply a simple linear extrapolation scheme to estimate intermediate features:
\begin{equation}
\begin{aligned}
    \mathcal{G}_{t-k}^l &= \mathcal{G}_t^l + \frac{\Delta \mathcal{G}_t^{l}}{M} \cdot k = \mathcal{G}_t^l + \frac{\mathcal{G}_t^l - \mathcal{G}_{t+M}^{l}}{M} \cdot k,
    \\
    \mathcal{H}_{t-k}^l &= \mathcal{H}_t^l + \frac{\Delta \mathcal{H}_t^{l}}{M} \cdot k  = \mathcal{H}_t^l + \frac{\mathcal{H}_t^l - \mathcal{H}_{t+M}^{l}}{M} \cdot k.
\end{aligned}
\end{equation}

By directly estimating intermediate features $\mathcal{G}_{t-k}^{l}$ and $\mathcal{H}_{t-k}^{l}$ in Eq.\,(\ref{eq:2}), the unified $z_{t-k}^{l}$ and text tokens $\tau_{t-k}^{l}$ can be updated through simple residual addition, thereby eliminating redundant computations of $\operatorname{FullAttn}$ and $\operatorname{FFN}$.

%% file: sec/4_exp.tex
\section{Experiments}
\subsection{Experimental settings}

\noindent
\textbf{Implementation Details.}
\method build upon Aether~\cite{Aether}, a recent world model framework fine-tuned from CogVideoX-I2V~\cite{CogvideoX}.
\method can integrate seamlessly into the inference of Aether as a plug-and-play module, without additional training.
Following the experimental settings of Aether, we evaluate our model on two tasks: \textit{Reconstruction} and \textit{Visual Planning}.
The hyper-parameter settings are as follows: $p_h=2$, $p_w=3$, $M=2,3$, $K=20$, $R=0.2$.
We set $T=30$ for the reconstruction task and $T=50$ for the visual planning task.
All experiments are executed on NVIDIA A100-80G GPUs.

\noindent
\textbf{Datasets Details.}
%
For reconstruction task, we evaluate our model on the Sintel~\cite{Sintel} dataset, which provides ground-truth depth and camera pose for accurate evaluation of reconstruction.
For visual planning task, we follow the experimental setup of Aether~\cite{Aether} and construct a custom validation set.
Specifically, we manually select over $111$ scenes from the RealEstate10K dataset, which cover both indoor and outdoor perspective transitions.
For each video clip, the first and last frames are used as input conditions to to assess visual planning performance.

\begin{figure*}[!t]
\centering
\includegraphics[width=0.98\textwidth]{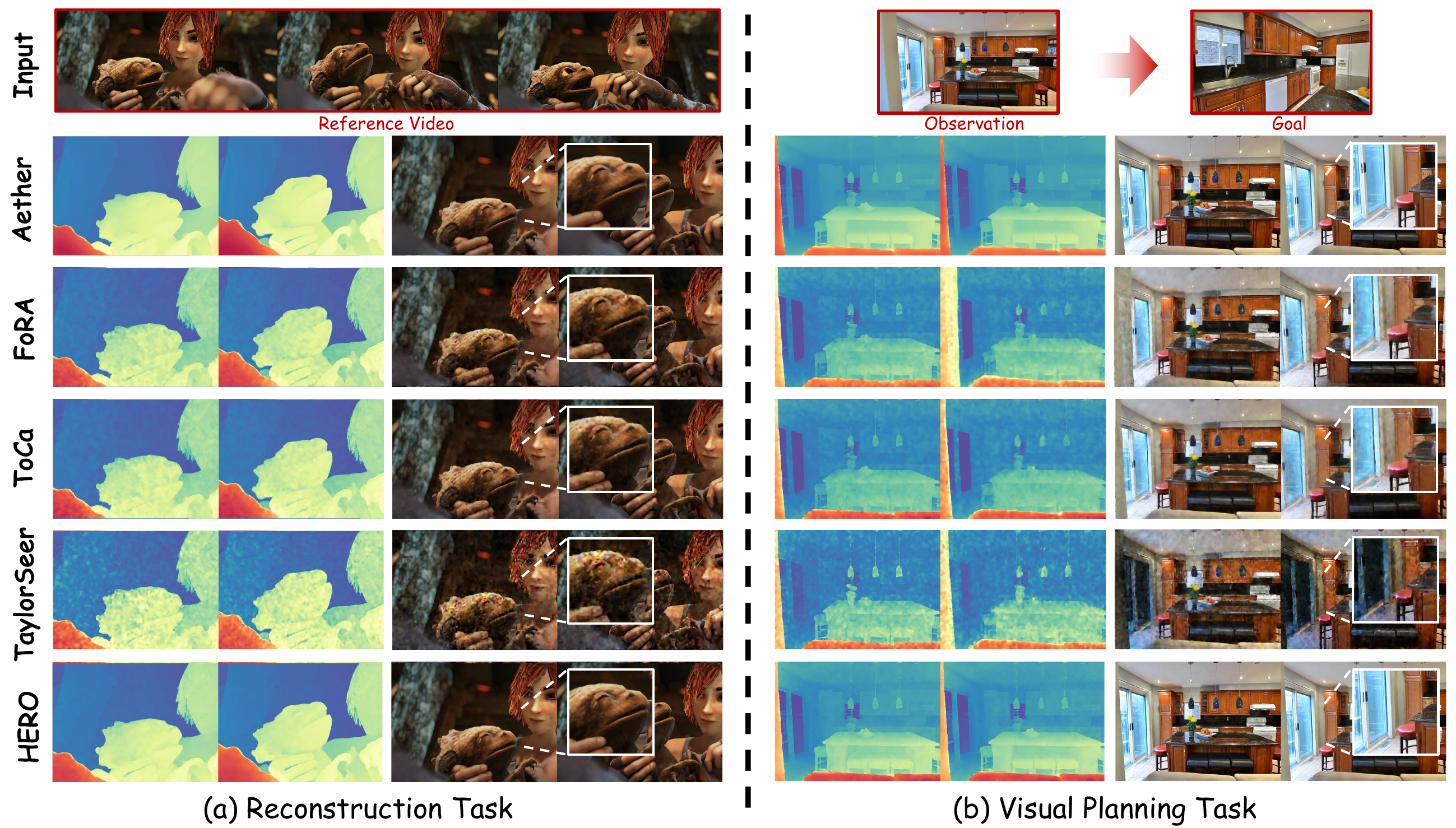} %
\caption{
Qualitative comparison with exsiting approaches in the (a) reconstruction task and the (b) visual planning task.
}
\label{fig:qualitative_comparison}
\end{figure*}

\subsection{Visual Planning Task}

\noindent
\textbf{Metrics.}
%
We systematically evaluate the generated videos using the same settings as Aether~\cite{Aether}.
Following the VBench~\cite{VBench} protocol, we quantitatively assess subject consistency, background consistency, motion smoothness, dynamic degree, aesthetic quality, and imaging quality.
As in the reconstruction task, we also report latency(s), flops(T) and speed to evaluate efficiency.

\noindent
\textbf{Baselines.}
%
We select CogVideoX-I2V-5B~\cite{CogvideoX}, the video generator, and Aether~\cite{Aether}, the world model, as our reference benchmarks.
Similar to reconstruction tasks, since no methods specifically accelerate world models, we compare our approach with three recent diffusion model acceleration methods: FoRA~\cite{FoRA}, ToCa~\cite{ToCa}, and TaylorSeer~\cite{TaylorSeer}, to evaluate the superiority of our method.

\noindent
\textbf{Experimental Analysis.}
In visual planning task, our \method delivers superior trade-off between performance and efficiency in qualitative and quantitative comparisons.
In Table\,\ref{tab:eval_video}, existing acceleration methods incur varying degrees of degradation on VBench.
Notably, TaylorSeer~\cite{TaylorSeer} and FoRA~\cite{FoRA} achieve notable speedups while sacrificing substantial performance compared to the original Aether~\cite{Aether}.
In contrast, \method achieves a 1.73$\times$ speedup with minimal performance drop.
In Figure\,\ref{fig:qualitative_comparison}, directly applying existing acceleration methods to world models results in degraded visual quality in visual planning tasks, manifesting as blurry and temporally inconsistent video frames and depth maps.
In contrast, our \method maintains high-quality visual planning outputs with more stable and coherent frame sequences.

\begin{table*}[!t]
\centering
\resizebox{\linewidth}{!}{  
\begin{tabular}{lccccccccc}
\toprule
Settings($K=15$) & Subj. Cons. $\uparrow$ & Back. Cons. $\uparrow$ & Mot. Smooth. $\uparrow$ & Dyn. Deg. $\uparrow$ & Aest. Qual. $\uparrow$ & Img. Qual. $\uparrow$ & Latency(s) $\downarrow$ & FLOPs(T) $\downarrow$ & Speed $\uparrow$ \\
\midrule
$R=0.8$           & 93.21 & 94.69 & 97.86 & 90.09 & 49.36 & 64.11 & 133.56 & 9816.90 & 1.46$\times$ \\
$R=0.7$           & 93.20 & 94.71 & 97.86 & 90.09 & 49.35 & 64.12 & 128.37 & 9404.28 & 1.52$\times$ \\
$R=0.5$           & 93.21 & 94.73 & 97.85 & 90.09 & 49.35 & 64.11 & 123.14 & 8991.43 & 1.59$\times$ \\
$R=0.3$           & 93.21 & 94.75 & 97.86 & 90.09 & 49.36 & 64.10 & 117.95 & 8578.81 & 1.67$\times$ \\
\rowcolor{gray!25}
$R=0.2$           & 93.21 & 94.72 & 97.86 & 90.09 & 49.35 & 64.12 & 112.73 & 8166.19 & 1.75$\times$ \\
\bottomrule
\end{tabular}
}
\caption{
Ablation studies on sample ratio $R$. It highlights the spatio-temporal redundancy in video content.
}
\label{table:ablation_R}
\end{table*}

\begin{table*}[!t]
\centering
\resizebox{\linewidth}{!}{  
\begin{tabular}{lccccccccc}
\toprule
Settings($R=0.2$) & Subj. Cons. $\uparrow$ & Back. Cons. $\uparrow$ & Mot. Smooth. $\uparrow$ & Dyn. Deg. $\uparrow$ & Aest. Qual. $\uparrow$ & Img. Qual. $\uparrow$ & Latency(s) $\downarrow$ & FLOPs(T) $\downarrow$ & Speed $\uparrow$ \\
\midrule
$K=5$      & 90.01 & 93.71 & 97.31 & 89.19 & 48.06 & 59.06 & 109.89 & 7908.13 & 1.81$\times$ \\
$K=10$     & 90.05 & 93.65 & 93.33 & 90.09 & 48.13 & 61.10 & 111.45 & 8037.16 & 1.78$\times$ \\
$K=15$     & 93.21 & 94.72 & 97.86 & 90.09 & 49.35 & 64.12 & 112.73 & 8166.19 & 1.75$\times$ \\
\rowcolor{gray!25}
$K=20$     & 93.42 & 95.02 & 97.90 & 90.09 & 49.34 & 64.21 & 113.98 & 8295.22 & 1.73$\times$ \\
$K=25$     & 93.40 & 95.02 & 97.92 & 90.09 & 49.22 & 64.02 & 115.32 & 8424.25 & 1.70$\times$ \\
$K=30$     & 93.44 & 95.11 & 97.97 & 90.09 & 49.24 & 63.83 & 117.02 & 8553.51 & 1.67$\times$ \\
$K=35$     & 93.44 & 95.12 & 97.97 & 90.09 & 49.24 & 64.06 & 118.87 & 8682.54 & 1.65$\times$ \\
\bottomrule
\end{tabular}
}
\caption{
Ablation on threshold $K$.
Smaller $K$ triggers more extrapolation errors in shallow features, leading to degraded results.
}
\label{table:ablation_K}
\end{table*}

\subsection{Reconstruction Task}

\noindent
\textbf{Metrics.}
In the reconstruction task, we evaluate model performance on two settings: depth estimation and camera pose estimation, following the same protocol as Aether~\cite{Aether}.
For depth estimation, we compare predicted and ground-truth depths frame by frame, with metrics including absolute relative error (Abs Rel), $\delta < 1.25$ (ratio of predictions within $1.25\,\times$ of the ground truth) and $\delta < 1.25^2$ as metrics.
For camera pose estimation, we assess per-frame consistency with ground-truth poses.
After Sim(3) alignment, we report absolute translation error (ATE), relative translation error (RPE Trans), and relative rotation error (RPE Rot).
Additionally, following prior works~\cite{FoRA,ToCa}, we report latency(s), flops(T) and speed to evaluate efficiency.

\noindent
\textbf{Baseline.}
We chose two reconstruction-based methods: DUST3R~\cite{DUST3R} and CUT3R~\cite{CUST3R}, along with the world model Aether~\cite{Aether} as our reference benchmarks.
Additionally, since no specific methods exist for world model acceleration, we selected three diffusion model acceleration methods: FoRA~\cite{FoRA}, ToCa~\cite{ToCa}, and TaylorSeer~\cite{TaylorSeer} to highlight the superiority of our \method.

\noindent
\textbf{Experimental Analysis.}
In reconstruction task, our \method also achieves a better trade-off between performance and efficiency in qualitative and quantitative comparisons.
As illustrated in Table\,\ref{tab:eval_recons}, existing acceleration method TaylorSeer~\cite{TaylorSeer} offers notable inference speedup but suffers from significant performance degradation, with an $\delta<1.25$ of 0.324 in video depth estimation and an RPE translation error of 1.105 in camera pose estimation.
In contrast, our \method achieves a 1.65$\times$ speedup while maintaining higher accuracy, achieving an $\delta<1.25$ of 0.502 and an RPE translation error of 0.816.
As shown in Figure\,\ref{fig:qualitative_comparison}, directly applying existing acceleration methods to world models degrades visual quality in reconstruction tasks, resulting in distortions, blurriness, and artifacts in video frames and depth maps.
In contrast, our \method preserves high reconstruction quality with more faithful and consistent results.

\subsection{Ablation Studies}
In this section, We first perform ablation studies on the sampling ratio $R$, then investigate the threshold parameter $K$, which defines the boundary between shallow and deep layers.
All experiments are conducted on visual planning task.

\noindent
\textbf{Ablation on Sample Ratio $R$.}
Recall that in the patch-wise refresh strategy, each divided patch is sampled with a certain ratio $R$.
We first coarsely set the threshold parameter $K=15$, and then proceed to determine the sampling ratio $R$.
As shown in Table\,\ref{table:ablation_R}, decreasing $R$ improves inference speed with negligible impact on generation quality, which highlights the spatio-temporal redundancy in video content.


\noindent
\textbf{Ablation on Threshold $K$.}
Recall that in the first $K$ MMdiT layers, we apply patch-wise refresh to dynamically recompute tokens, thereby reducing errors and improving quality fidelity.
In the remaining layers, we use linear extrapolation to skip the computations of $\operatorname{FullAttn}$ and $\operatorname{FFN}$, thus enabling faster inference.
With the sampling ratio fixed at $R=0.2$, we then investigate the optimal threshold $R$, as shown in Table\,\ref{table:ablation_K}.
As $K$ decreases, linear extrapolation is used more frequently, speeding up inference but introducing extrapolation errors in shallow features, which reduces generation quality.
As $K$ increases, patch-wise refresh is used more often, slightly improving quality but introducing additional refresh computations, which limit inference speed.
This creates a trade-off between quality and speed.

%% file: sec/5_conclusion.tex

\section{Conclusion}
In this paper, we present \method, a training-free hierarchical acceleration framework for efficient inference in world models.
Due to the nature of multi-modal, world models exhibit feature coupling, with shallow layers showing high temporal variability and deeper layers producing more stable representations.
To address this, \method adopts a hierarchical strategy leveraging inherent characteristics:
(i) In shallow layers, a patch-wise refresh mechanism dynamically selects tokens for recomputation.
With patch-wise sampling and frequency-aware tracking, it avoids additional metric computation while remaining compatible with FlashAttention.
(ii) In deeper layers, a linear extrapolation scheme estimates intermediate features, bypassing redundant computations in attention and feed-forward networks.
Extensive experiments show that \method outperforms existing acceleration baselines in both efficiency and accuracy.

%% file: sec/X_suppl.tex
\appendix

\section{Additional Implementation Details}
We provide more detailed implementation settings to facilitate reproducibility.
As described in the main paper, we adopt Aether~\cite{Aether}, a recent state-of-the-art world model, as our foundational framework.
It is worth noting that for the reconstruction task, the video resolutions in the Sintel~\cite{Sintel} dataset are not fixed and typically do not match the default resolution of 480$\times$720 supported by Aether.
Directly resizing the input videos to 480$\times$720 leads to a noticeable drop in performance, as also reported in the original Aether paper. To address this issue, we follow the experimental setup in Aether by applying a sliding window of size 480$\times$720 to process each video.
The model reconstructs each patch individually, and the final result is obtained by averaging predictions in the overlapping regions.
To ensure reproducibility and a fair comparison, we strictly follow the evaluation protocol provided in the official Aether codebase without any modifications.

\section{Additional Qualitative Comparison}
\begin{figure*}[!t]
\centering
\includegraphics[width=0.98\textwidth]{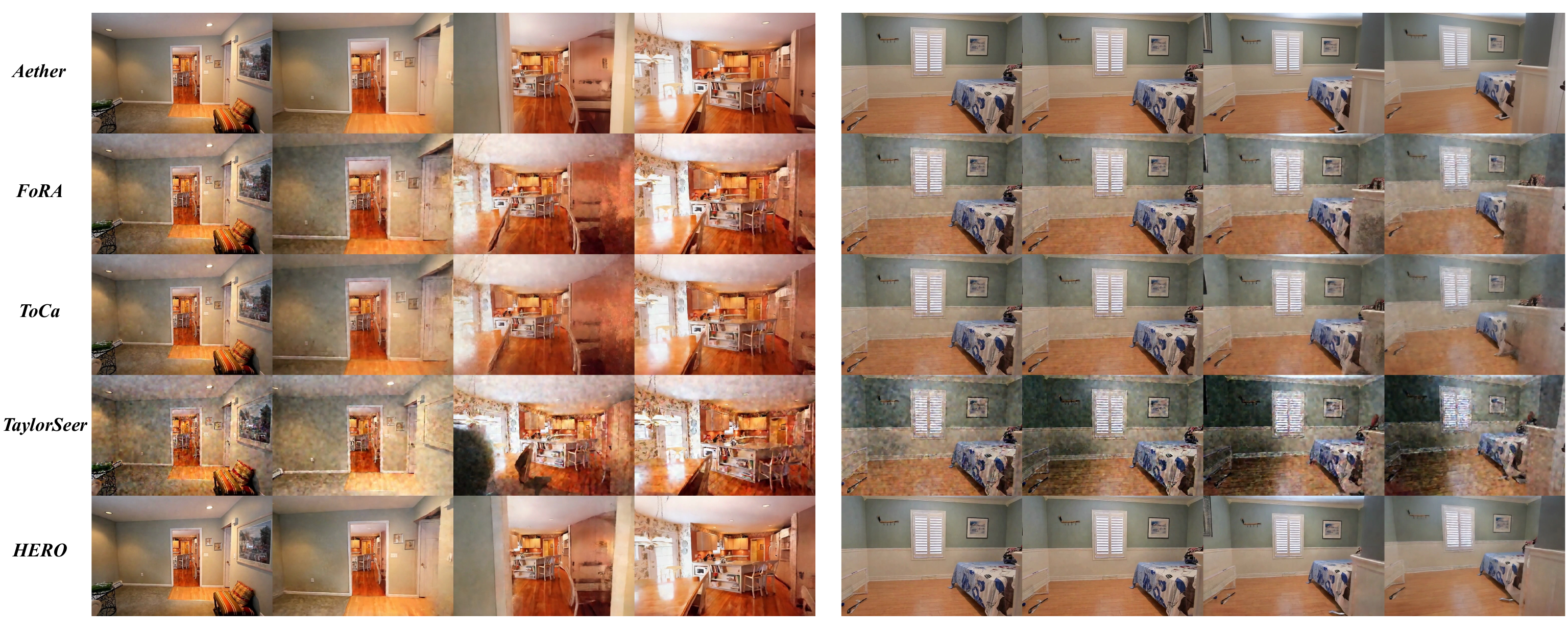} %
\caption{
Additional qualitative comparison with existing approaches in the visual planning task.
}
\label{fig:additional_comparison}
\end{figure*}

\begin{figure*}[!t]
\centering
\includegraphics[width=0.98\textwidth]{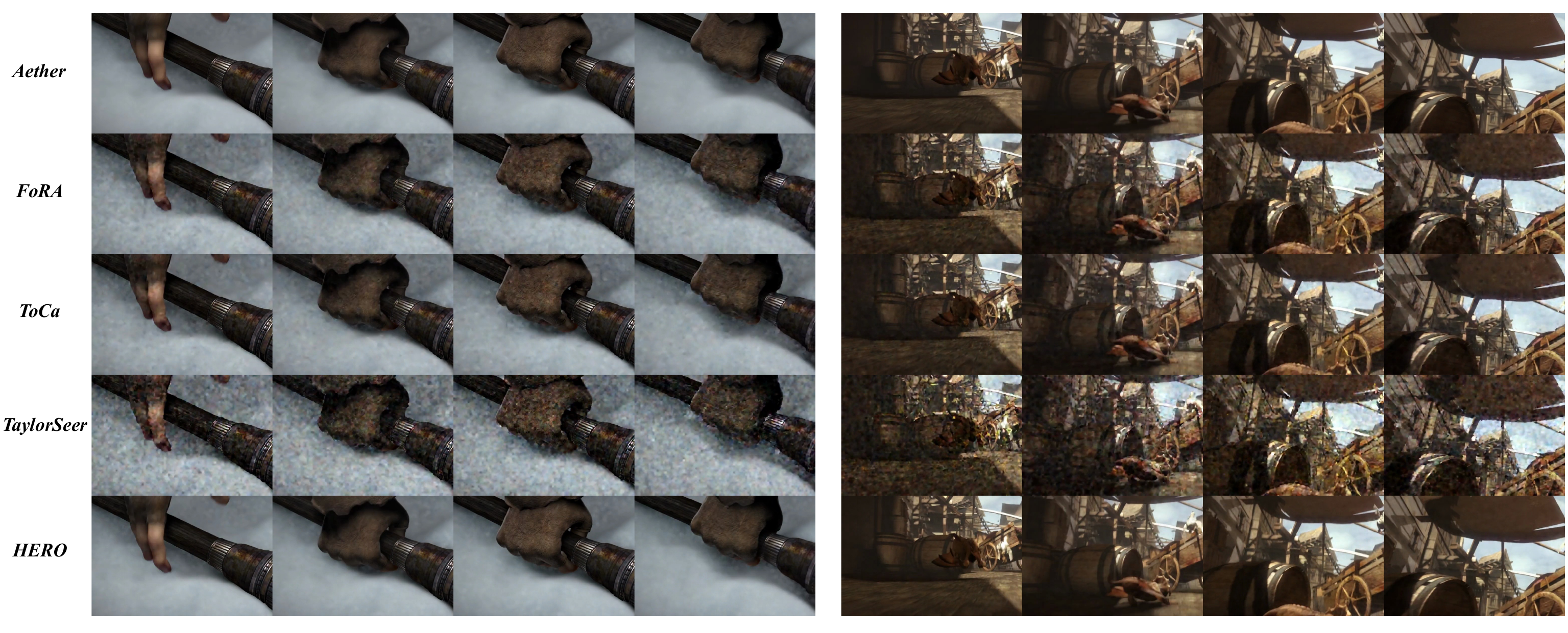} %
\caption{
Additional qualitative comparison with existing approaches in the reconstruction task.
}
\label{fig:additional_comparison2}
\end{figure*}

We provide additional qualitative comparisons between our \method and several diffusion acceleration baselines, including FoRA~\cite{FoRA}, ToCa~\cite{ToCa}, TaylorSeer~\cite{TaylorSeer}, and the original Aether~\cite{Aether}, on both the reconstruction and visual planning tasks. Representative results are shown in Figure\,\ref{fig:additional_comparison} and Figure\,\ref{fig:additional_comparison2}.
It is evident that directly applying existing diffusion acceleration methods within the world model framework, such as Aether, leads to degraded generation quality, including both video content and depth maps. In contrast, our \method achieves efficient inference without compromising performance.
These results support our key insight: world models exhibit hierarchical representations due to multi-modal coupling.
Uniform acceleration across layers ignores this structure, leading to degradation. In contrast, \method uses a layer-wise strategy aligned with this hierarchy, achieving acceleration without sacrificing fidelity.

\section{Additional Visualization Results}
\begin{figure*}[!t]
\centering
\includegraphics[width=0.98\textwidth]{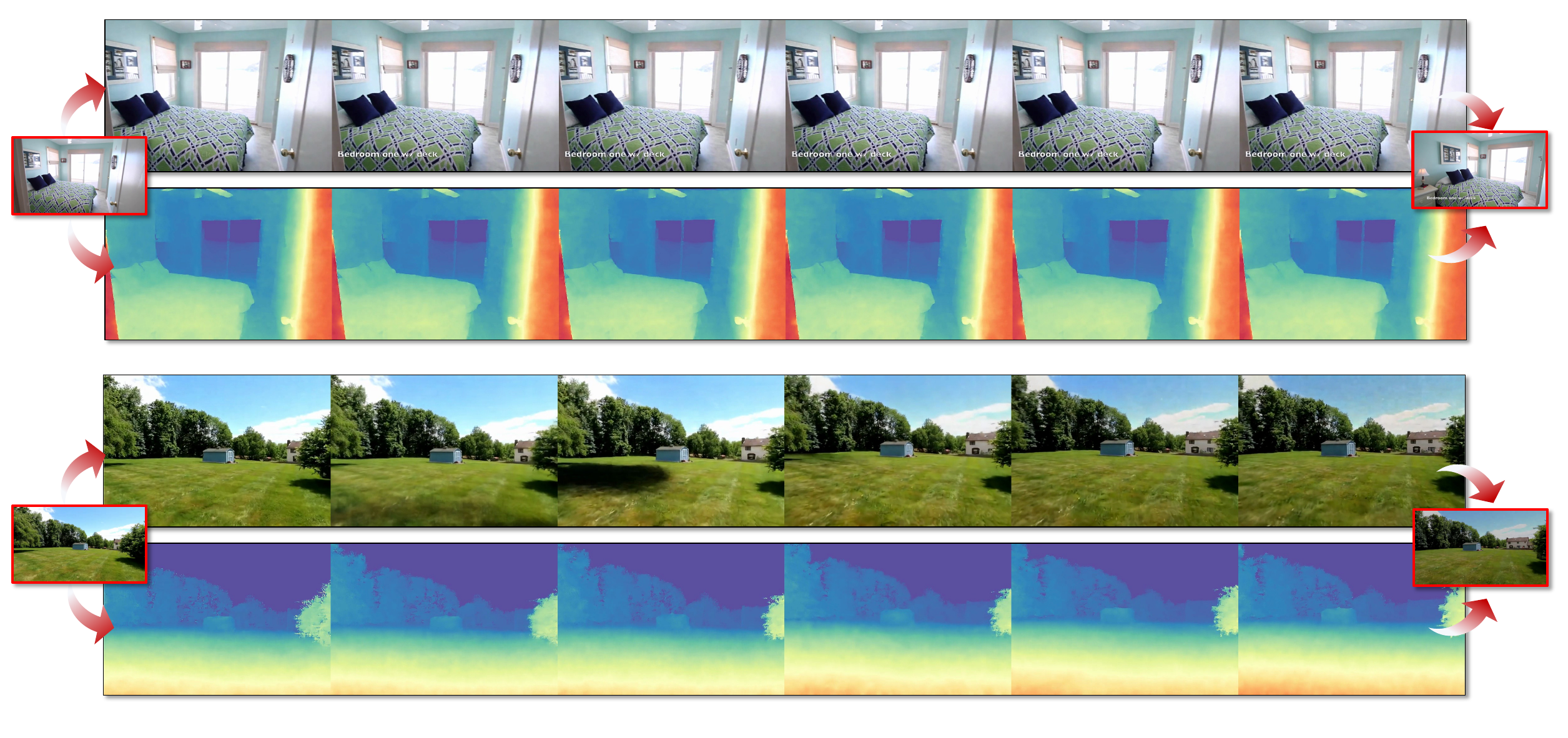} %
\caption{
Additional visualization results of our \method in the visual planning task.
}
\label{fig:gallery}
\end{figure*}

\begin{figure*}[!t]
\centering
\includegraphics[width=0.98\textwidth]{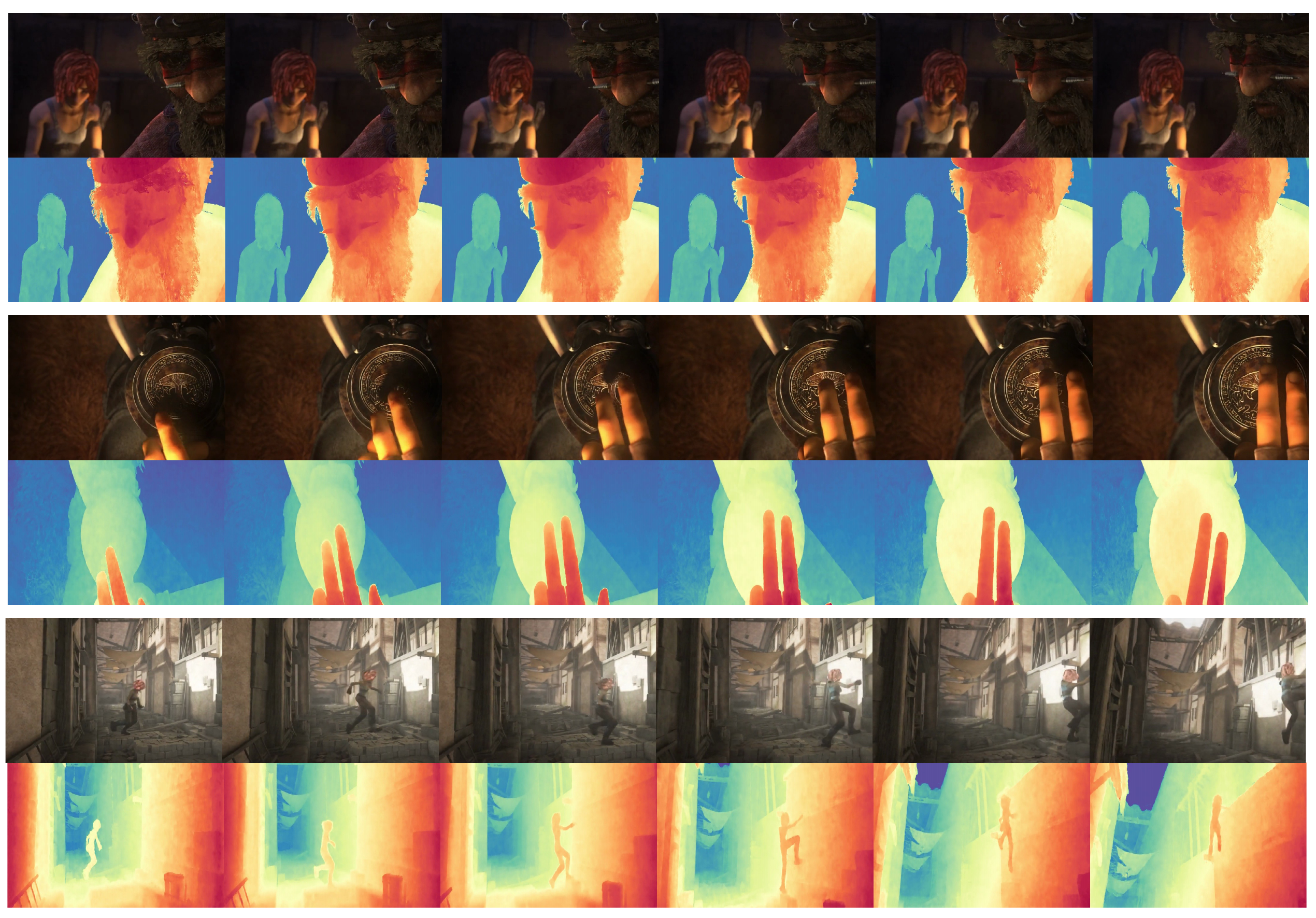} %
\caption{
Additional visualization results of our \method in the reconstruction task.
}
\label{fig:gallery2}
\end{figure*}

To further demonstrate the advantages of our \method, we present additional visualization results on both the reconstruction and visual planning tasks. Detailed examples are shown in Figure\,\ref{fig:gallery} and Figure\,\ref{fig:gallery2}.
Our \method accelerates world model inference while maintaining performance with minimal degradation, further demonstrating the effectiveness of its hierarchical  acceleration strategy in handling the coupled feature structure inherent in world models.

